\documentclass{article}

\usepackage[english]{babel}
\usepackage{authblk}

\usepackage[letterpaper,top=2cm,bottom=2cm,left=3cm,right=3cm,marginparwidth=1.75cm]{geometry}

\usepackage{amsmath}
\usepackage{graphicx}
\usepackage[colorlinks=true, allcolors=blue]{hyperref}
\usepackage{amsmath}
\usepackage{amsfonts}
\usepackage{booktabs} 
\usepackage{longtable} 
\usepackage[utf8]{inputenc}
\usepackage[T1]{fontenc}
\usepackage{array}
\usepackage{booktabs}

\newcommand{\mpar}{\medskip\noindent}

\title{Feasibility of machine learning-based rice yield prediction in India at the district level using climate reanalysis data}
\author{Djavan De Clercq, Adam Mahdi}
\affil{University of Oxford}
\date{}

\begin{document}
\maketitle

\begin{abstract}
Yield forecasting, the science of predicting agricultural productivity before the crop harvest occurs, helps a wide range of stakeholders make better decisions around agricultural planning. This study aims to investigate whether machine learning-based yield prediction models can capably predict Kharif season rice yields at the district level in India several months before the rice harvest takes place. The methodology involved training 19 machine learning models such as CatBoost, LightGBM, Orthogonal Matching Pursuit, and Extremely Randomized Trees on 20 years of climate, satellite, and rice yield data across 247 of India’s rice-producing districts. In addition to model-building, a dynamic dashboard was built understand how the reliability of rice yield predictions varies across districts. The results of the proof-of-concept machine learning pipeline demonstrated that rice yields can be predicted with a reasonable degree of accuracy, with out-of-sample R2, MAE, and MAPE performance of up to $0.82$, $0.29$, and $0.16$ respectively. These results outperformed test set performance reported in related literature on rice yield modeling in other contexts and countries. In addition, SHAP value analysis was conducted to infer both the importance and directional impact of the climate and remote sensing variables included in the model. Important features driving rice yields included temperature, soil water volume, and leaf area index. In particular, higher temperatures in August correlate with increased rice yields, particularly when the leaf area index in August is also high. Building on the results, a proof-of-concept dashboard was developed to allow users to easily explore which districts may experience a rise or fall in yield relative to the previous year. The dashboard show that the model may perform better in some regions than in others. For instance, the absolute percentage error for predicted versus actual yields ranged from an average of $7.1\%$ in districts in Uttarakhand to an average of $14.7\%$ in Uttar Pradesh. This study underscores the potential for policymakers to consider scaling and operationalizing machine learning approaches to rice yield prediction in the context of agricultural early warning systems to deliver timely crop yield forecasts on a rolling basis throughout the season, thereby equipping agricultural decision-makers with the ability to make informed choices on irrigation scheduling, fertilizer application, and harvest planning to optimize crop output and resource use. 
\end{abstract}

\section{Introduction}

\subsection{The societal implications of accurate crop yield forecasting in India}

Yield forecasting is the science of predicting agricultural productivity as measured by crop yield – the ratio of the total mass of the harvested product (such as rice) to the area used to cultivate the crop  – before the harvest takes place, typically a few months in advance \cite{Weiss2020Jan}. 

Pre-harvest prediction of crop yields is important in helping a wide range of stakeholders make better decisions around agricultural planning. For farmers, accurate crop yield forecasts can facilitate decision-making around what to grow and when to grow it \cite{vanKlompenburg2020Oct}. In addition, near real-time monitoring of crop growth can inform the use of preventive measures such as irrigation and fertilization to boost agricultural productivity where needed \cite{Sagan2021Apr}. For governments, yield prediction is relevant to the formulation of policies related to national food security, such as pricing policies for domestic markets, and policy decisions on the import and export of different crops \cite{Li2007Oct}.

Accurate crop yield forecasting may also enable better design of insurance products that mitigate climate risks and stabilize farmer incomes \cite{Wang2021Dec}. Weather-based crop insurance, for instance, uses a weather index such as total precipitation to determine payments to farmers, meaning that insurance companies do not need to visit farmers to assess damages and arbitrate claims. Rather, if the weather reaches a certain threshold, rapid automatic payments can be distributed to farmers, who avoid the need to sell assets to survive due to adverse climate events \cite{wen2019application}.

The need for accurate information on crop yields is particularly important in countries like India, where the agricultural sector provides livelihoods for hundreds of millions of farmers, with $70\%$ of rural households depending on agriculture for their main source of income. One of India’s major staple crops is rice, which contributes to 30\% of calories consumed in India and is a key export commodity for the country \cite{nayak2022rice}. India cultivates rice on about 45 million hectares of land, with a total production of 178 million metric tonnes in 2020 \cite{fao_india_2018}. In addition, the distribution of monsoon rainfall, which is a major source of water for rice cultivation, has become erratic in recent years due to climate variability \cite{ishfaq2020alternate, sattar2021modelling}. In such contexts, crop yield predictions may be able to supplement agricultural early warning systems (AEWSs) that give advanced notice of potential risks to crop productivity, enabling preemptive action in affected areas. Previous research has shown that current agricultural monitoring systems lack robust crop yield and crop production forecasts, as well as the operationalisation of such methods at scale \cite{fritz2019comparison}. 

\subsection{Overview of approaches and variables used to model crop yields}

Crop yield prediction is a challenging problem in precision agriculture, as final yields depend on a variety of factors such as weather, climate, soil, seed type, and agronomic practices such as irrigation and fertilizer use \cite{xu2019design}. 

This complexity is evident from the variety of variables included and methods applied in the growing body of literature on crop yield forecasting. For example, recent examples in literature involving deep learning approaches include corn and soybean yield forecasting in the US based on convolutional neural networks (CNNs) and recurrent neural networks (RNNs) \cite{khaki2020cnn}, soybean yield forecasting in Argentina based on deep transfer learning \cite{wang2018deep}, and vineyard grape yield estimations based on CNNs \cite{Kamangir2024Jan}. Recent examples based on machine learning approaches include sugarcane yield prediction using random forests \cite{charoen2019sugarcane}, prediction of wheat, barley, and canola yields in Western Australia using random forest \cite{filippi2019approach}, yield forecasting of spring maize in Pakistan based on LASSO regression and support vector machine \cite{ahmad2018yield}, and Jojoba yield prediction in Israel based on gradient boosted regression trees \cite{goldstein2018applying}. Other examples of the machine learning approaches that have been applied to yield prediction have been summarized in a systematic literature review, which also analyzed the variables most frequently included in crop yield prediction studies. Across 50 studies between 2008 and 2019, features used as predictors of yield have included temperature, soil type, rainfall, humidity, pH-value, NDVI, wind speed, and more \cite{vanKlompenburg2020Oct}. 

For studies specific to rice, the staple crop of over half the world’s population \cite{chaurasiya2022layering}, a number of approaches have been applied to yield forecasting in recent years. Recent examples include rice yield prediction for 81 counties in southern China based on recurrent neural networks \cite{chu2020end}; application of the ecological distance algorithm to model rice yields \cite{tian2020yield}; field and county-level rice yield prediction based on synthetic aperture radar (SAR), optical and meteorological data \cite{yu2023improved}; random forest yield prediction based on high-resolution imagery collected from unmanned ariel vehicles (UAVs) \cite{wan2020grain}; simulation of yields using the Cropping System Model-CERES-RICE \cite{jha2019using}; pixel-scale rice yield prediction in South Korea based on a combination of deep learning and crop models \cite{jeong2022predicting}; rice yield estimation at 500m spatial resolution based on gradient boosted regression and vegetation indices derived from the Moderate Resolution Imaging Spectroradiometer (MODIS) \cite{arumugam2021remote}; and rice paddy yield prediction using sentinel-based optical and SAR data in India based on random forest \cite{ranjan2019paddy}. To the authors’ knowledge, there has been less research on rice prediction at the district-level in India. 

\subsection{Research contributions}

This study marks an advancement in the field of rice yield prediction in India by building a proof-of-concept approach capable of predicting rice yields at the district level for 247 rice-producing districts across India. 

First, a novel combination of data sources is used to predict Indian rice yields. These include data from ERA5, a climate re-analysis product developed by the European Centre for Medium Range Weather Forecasts (ECMWF), which combines observations with modelled data to provide hourly data on atmospheric, land-surface, and sea-state parameters globally \cite{gomez2021regional}. Vegetation data was derived from the MODIS sensors on-board NASA’s TERRA and AQUA satellites, and a cropland mask (CROPGRIDS) was used to filter earth observation data according to where rice is grown at a pixel level. Collectively, these data enhance the model’s ability to capture the intricate effects of climate variables and vegetation health indices on crop yields. 

Secondly, the study creates a spatially matched dataset for rice crop yields in India, tackling a common problem in yield prediction research: a lack of training datasets where yield data is accurately aligned with specific geographic locations. Typically, discrepancies between the names and boundaries in geographic datasets (like shapefiles) and official agriculture statistics can hinder the effective use of data in research. Fuzzy matching algorithms were used to align Indian shapefiles (which may have variations in district names) with official rice yield data from the Indian government to produce a dataset where yield information is precisely matched to its geographic location. This matching is important for researchers aiming to spatially aggregate earth observation data in a manner consistent with available yield data. Making this matched dataset open source specifically aids rice-related research in India, enabling more accurate earth observation studies by providing a reliable foundation for correlating satellite data with actual agricultural outputs.

Thirdly, the study provides a new benchmark for district-level rice yield prediction in India based on predictions made exploring the predictive effectiveness of 19 machine learning models such as LightGBM, Bayesian Ridge Regression, and others. This methodological approach facilitates the identification of the most effective models for rice yield prediction at the district level. Previous literature on crop yield prediction in India has largely focused on using a narrower range of algorithms such as random forest or support vector machine \cite{vanKlompenburg2020Oct}. A detailed evaluation of model performance is provided, coupled with the interpretability provided by SHAP values. 

Fourth, beyond model development, an interactive dashboard tool was developed, not only to visualize the yield predictions across each of India's districts, but also to allow for detailed diagnosis of model performance. It serves as a practical tool for model evaluation, offering insights into regional performance variations and facilitating the identification of areas where predictions may be improved.

Overall, this study offers evidence that scalable crop yield prediction models have the potential to be integrated into agricultural early warning systems in India, which, as noted in previous research, currently lack such forecasting capabilities and the means to operationalize these methods at a scale that contributes towards more resilient agricultural practices and food security planning \cite{fritz2019comparison}.

\section{Methods}

\subsection{Study region and brief overview}

In this study, climate and remote sensing data were used as predictors to model rice yields for the kharif season (wet summer monsoon season) from 2001 to 2020 at the district level in India (India consists of 36 states and 684 districts). In India, more than half of the annual rice crop is grown during kharif \cite{auffhammer2012climate}, a season which is characterized by high temperature, high humidity, and medium to high rainfall \cite{sharma2019field}. Kharif season rice is typically sowed between the start of June to the end of August and harvested between the end of September to early January, depending on the region. During the 2019-2020 season, harvesting of Kharif rice was completed in February 2020. 

The methods applied in this study can be summarized as follows. Firstly, 20 years of data on climate, vegetation, and rice yields were ingested programmatically via API from various sources. Second, climate and vegetation data were pre-processed and aggregated to the same level as historical rice yields prior to development of machine learning models. Third, a range of models including Bayesian ridge regression and LightGBM were trained, evaluated, and interpreted. Fourth, an interactive dashboard tool was developed to visualize the results of the algorithms to provide a spatially explicit view of model results and potential modelling errors. Lastly, the implications of the modelling results were discussed with regards to their inclusion in agricultural early warning systems. 

The methods outlined in this research are fully reproducible. All data, python code, and dashboards can be found on GitHub.

\subsection{Data collection}

\begin{figure}
    \centering
    \includegraphics[width=1\linewidth]{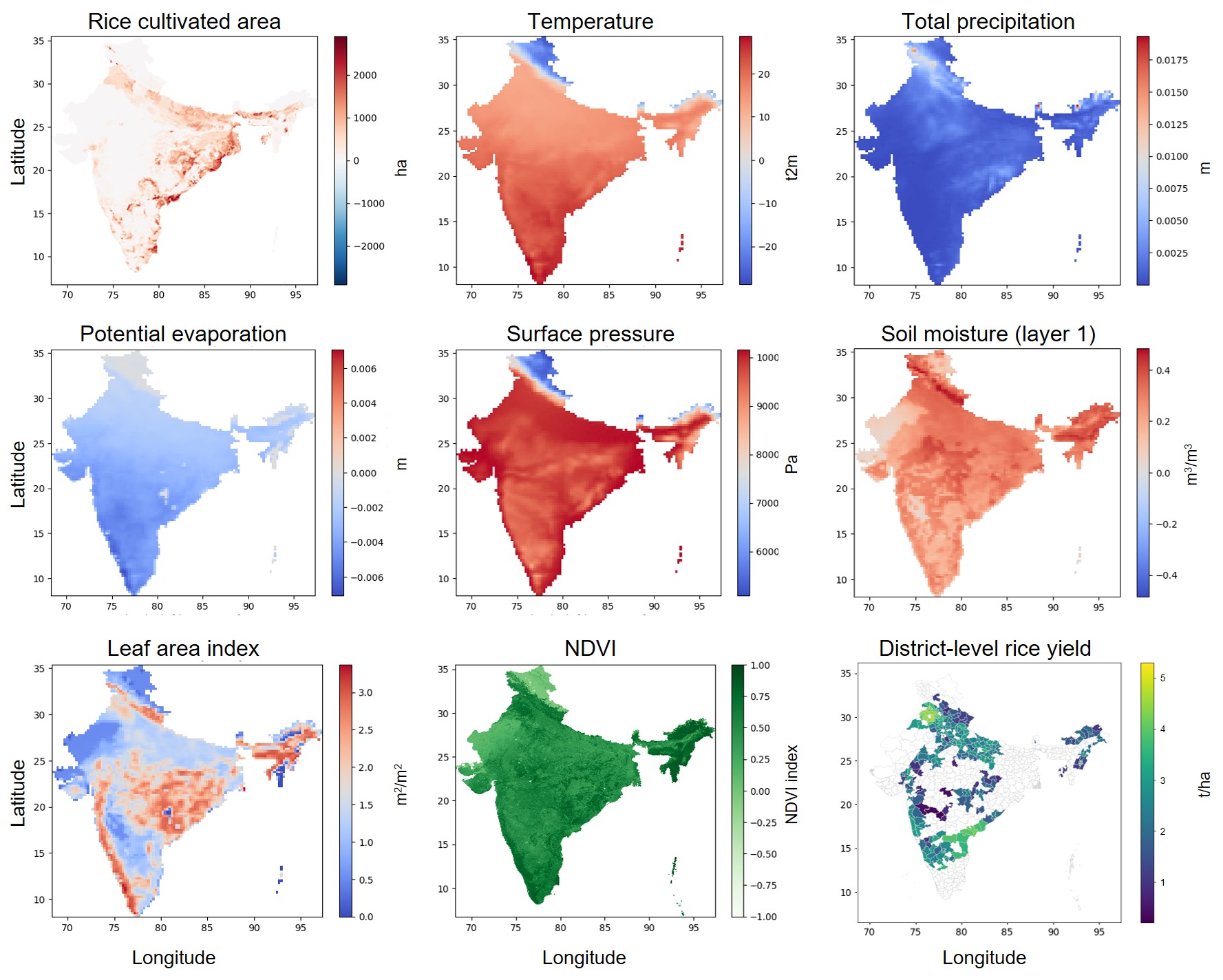}
    \caption{Overview of the geospatial data used in this research. The panels show, respectively: cultivated rice area in India; snapshots of temperature, total precipitation, potential evaporation, surface pressure, soil moisture, and leaf area index from the ECMFW (average values of January 2022); NDVI data from NASA’s MODIS (average values of January 2022); and district-level rice yield data from India’s Ministry of Agriculture and Farmers Welfare (in 2020).}
    \label{Figure 1}
\end{figure}

As shown in Table \ref{Table 1}, a range of datasets were used in the rice yield  modelling, including climate reanalysis data, remote sensing data, district-wise historical rice yield data, and cropland masks. A visual overview of the geospatial data used in this research are also provided in Figure \ref{Figure 1}. 

\mpar
\textbf{Climate reanalysis data.} Daily climate reanalysis data on temperature, potential evaporation, surface  pressure, leaf area index, total precipitation, and soil water content was obtained from ERA5 data from the European Center for Medium-Range Weather Forecasts (ECMWF), which provides global estimates of surface and atmospheric parameters since 1950 at a resolution of approximately 30*30 km \cite{hersbach2020era5}. Climate reanalysis data, which are often freely available, provide temporally and spatially homogenous data \cite{urban2021evaluation}, which makes them suitable for applications such as crop yield prediction in contexts where in-situ weather station measurements are inadequate or incomplete. In addition, weather stations vary in their accuracy and generally record a limited number of variables, such as rainfall, temperature, pressure, and wind speed; variables that are more technically demanding to measure, such as humidity and solar radiation, may be lacking \cite{colston2018evaluating}. 

The climate variables used in this study were selected due to their influence on rice yields. An extensive body of research has shown that rice growth is affected by factors such as soil water content \cite{chakraborty2021assessing}, temperature \cite{jia2019effects}, potential evaporation (as a proxy for transpirational demand) \cite{kuwagata2021hydrometeorology}, surface pressure \cite{tang2010effects}, and precipitation \cite{poddar2021effect}. These variables can impact rice development across phenological stages. Past case studies in India have shown the approximate number of days taken for each stage: the sowing to tillering phase (P1) can range from 30 to 60 days, and the rate of tillering tends to increase under higher temperatures; the tillering to panicle initiation phase (P2) can range between 42 to 49 days; the panicle initiation to flowering phase (P3) can range from 12 to 28 days; the flowering to milk phase (P4) can range from 7 to 20 days; and the mil to physiological maturity phase (P5) can range from 17 to 31 days \cite{bal2023critical}. One study showed reported a linear relationship between the days taken from sowing to flowering and average air temperature \cite{alvarado2002influence}. 

\mpar
\textbf{Remote sensing data.} Normalized Difference Vegetation Index (NDVI) is a dimensionless index that describes the difference between visible and near-infrared reflectance of vegetation cover, and can be used to estimate the density of green on an area of land \cite{schinasi2018modification}. To determine the density of green on a patch of land, the wavelengths of visible and near-infrared sunlight reflected by the plants are observed. NDVI values range from -1 to +1; higher values of NDVI imply healthy and dense vegetation, whereas lower NDVI values indicate sparser vegetation. NDVI data was obtained from the Moderate Resolution Imaging Spectroradiometer (MODIS) onboard NASA’s Terra and Aqua satellites, due to their wide coverage and temporal resolution. There are several examples in academic literature of using NDVI to investigate the progress of crops, such as for wheat in Argentina \cite{lopresti2015relationship}, cereals in Europe \cite{panek2020analysis}, and rice in Vietnam \cite{son2014comparative}. NDVI data was masked using CROPGRIDS, a global, geo-referenced dataset providing information on areas for 173 crops circa the year 2020, at a resolution of $0.05^\circ$ ($\sim 5.55$ km at the equator) \cite{tang2023cropgrids}. 

\mpar
\textbf{Yield data.} District-level rice production and yield data from 1995 to 2021 for 367 districts were obtained from the APY dataset of the Directorate of Economics and Statistics in India’s Ministry of Agriculture and Farmers Welfare \cite{moa2021}. In this dataset, the year denotes the year in which the crop was harvested. For kharif season rice, the sowing is in the previous calendar year \cite{sonkar2019vulnerability}. The raw yield data were aligned with a shapefile representing Indian districts. This alignment was achieved using the FuzzyWuzzy library in Python, which employs the Levenshtein distance to address the challenge of slight spelling discrepancies in the district names between the yield data and the shapefile \cite{diwan2023ai}. 

Data used in this analysis were programmatically downloaded via API and automated Python scripts. ERA5 data was ingested via the CDS API, while NDVI data was ingested via USGS’ AppEEARS API.

\begin{table}
\small
\caption{A range of agronomically-relevant datasets were used as predictors of the target variable (district-level rice yield in India); rice areas masks were used to filter NDVI data by rice-growing area }
\vspace{10pt} 
\label{Table 1}
\begin{tabular}{>{\raggedright\arraybackslash}p{0.1\linewidth}>{\raggedright\arraybackslash}p{0.15\linewidth}>{\raggedright\arraybackslash}p{0.41\linewidth}>{\raggedright\arraybackslash}p{0.1\linewidth}>{\raggedright\arraybackslash}p{0.1\linewidth}}
\toprule
\textbf{Data type}& \textbf{Parameter}&  \textbf{Description}& \textbf{Unit}& \textbf{Source} \\ 
\midrule
Climate reanalysis& Potential evaporation (pev)&  A measure of the extent to which near-surface atmospheric conditions are conducive to the process of evaporation.& m& ECMFW (ERA5) \\
& 2m-temperature&  The temperature of air at 2m above the surface of land, sea or inland waters. 2m temperature is calculated by interpolating between the lowest model level and the Earth's surface, taking account of the atmospheric conditions.& K& ECMFW (ERA5) \\
& Total precipitation& The accumulated liquid and frozen water, comprising rain and snow, that falls to the Earth's surface. It is the sum of large-scale precipitation and convective precipitation.& m&ECMFW (ERA5) \\
& Leaf area index, low vegetation (LAI) & The surface area of one side of all the leaves found over an area of land for vegetation classified as 'low'.  'Low vegetation' consists of crops and mixed farming, irrigated crops, short grass, and more.& m\textsuperscript{2}-m\textsuperscript{-2}&ECMFW (ERA5) \\
& Total precipitation& The accumulated liquid and frozen water, comprising rain and snow, that falls to the Earth's surface. It is the sum of large-scale precipitation and convective precipitation.& m&ECMFW (ERA5) \\
 Climate reanalysis& Volumetric soil water (SWVL1) & The volume of water in soil layer 1 (0 - 7cm, the surface is at 0cm)& m\textsuperscript{3}-m\textsuperscript{-3}&ECMFW (ERA5) \\
 Remote Sensing& Normalized Difference Vegetation Index (NDVI) & A dimensionless index that describes the difference between visible and near-infrared reflectance of vegetation cover, and can be used to estimate the density of green on an area of land& (-)&NASA EOSDIS (AQUA MODIS) \\
Crop mask& Rice crop mask&  A comprehensive global, geo-referenced dataset providing information on areas for 173 crops circa the year 2020, at a resolution of 0.05° (~5.55 km at the equator).& ha& CROP \\
\bottomrule
\end{tabular}
\end{table}

\subsection{Data pre-processing}

Climate data from ERA5 in NetCDF format over a bounded area comprising India was clipped to the Indian country boundary. NDVI data from NASA’s AQUA MODIS satellite in NetCDF format were clipped to the Indian country boundary and masked with a rice cropland layer. 

Next, the climate variables and NDVI were aggregated to the district level based on zonal statistics. The vector geometry data for India’s ADM2 (district-level) boundaries which raster pixels were aggregated to were obtained from the Database of Global Administrative Areas (GADM) \cite{gadm2012}. District-level yield data from APY was then merged to the climate and remote sensing data aggregated at the district level to produce a spatially consistent geodataframe. Yield outliers beyond three standard deviations were removed as they were assumed not achievable at the district level in India \cite{arumugam2021remote}. 

Feature engineering was conducted to produce monthly averages for the climate and NDVI parameters for every month between May and November, corresponding to the full sowing and growing period for kharif rice \cite{arumugam2021remote}.  This process was repeated for all variables to produce a set of 52 features used as input for the modelling. The months selected for climate and NDVI feature aggregation were chosen to reflect the full range of rice growth stages, including the grain filing, vegetative, and reproductive stages \cite{zhou2023improved}. 

\subsection{Model development and interpretation}

This study developed and tested the performance of multiple rice yield prediction models based on a variety of machine learning models. These included LightGBM \cite{ke2017lightgbm}, an efficient and distributed gradient boosting framework that uses tree-based learning, Bayesian ridge regression \cite{shi2016bayesian, tipping2001sparse, mackay1992bayesian}, which has been recognized for its ability to deal with hierarchical data structures \cite{huang2010multilevel}, gradient boosting regression \cite{friedman2002stochastic}, random forest \cite{breiman2001random}, Huber regression \cite{huber2009robust}, decision tree regression \cite{hastie2009elements}, elastic net regression \cite{zou2005regularization}, AdaBoost \cite{hastie2009multi}, orthogonal matching pursuit \cite{rubinstein2008efficient}, and extremely randomized trees \cite{geurts2006extremely}. 

The models above were trained on district-level data for 2001 to 2018 (4,606 observations), and validated on out-of-sample test data for 2019 and 2020 (502 observations). The data was split in a manner that reflects how yield prediction models may be used in practice, avoiding random splits in favor of chronological splits to help ensure the model’s robustness to future, unseen data. This out-of-sample approach to testing regression models with temporal dependency has been shown to be more robust than cross-validation approaches tailored to time series problems \cite{cerqueira2020evaluating}.  

The top-performing models were evaluated based on three out-of-sample performance measures including R2, Mean Absolute Error (MAE), and Root Mean Square Error (RMSE). Also reported were Mean Squared Error (MSE), Mean Absolute Percentage Error (MAPE), and Root Mean Squared Logarithmic Error (RMSLE). In addition, model results were evaluated based on prediction error plots, residual plots, and spatio-temporal plots of prediction error to evaluate potential model bias (for example, better model performance for certain rice-growing regions of India). Lastly, Shapley Additive exPlanations (SHAP) were used to explore the impact of features on model output \cite{lundberg2017unified}. SHAP values are a model-independent methodology used for quantifying the significance of features in predictive modeling. The SHAP of feature $j$ for observation $\phi_j(\mathbf{x})$ is defined as
\begin{equation}
\phi_j(\mathbf{x}) = \sum_{S \subseteq \{1,\ldots,M\} \setminus \{j\}} \frac{|S|! \cdot (M - |S| - 1)!}{M!} \left( f(\mathbf{x}_{S \cup \{j\}}) - f(\mathbf{x}_S) \right)
\end{equation}
where $j$ is the feature evaluated, $M$ the total number of features, $S$ a subset of the full feature set \{1,\ldots,M\} that does not include the feature $j$, $\mathbf{x}_S$ a subset of features in $S$, and $f$ the model's prediction function \cite{lenaers2023exploring}.

\subsection{Computation}
Data ingestion, pre-processing, and modelling was conducted in a conda-based python environment with a diverse set of python libraries. Data processing and geospatial operations were carried out using python libraries including numpy, xarray, pandas, rasterio, rasterstat, and geopandas. Modelling and visualization was conducted using python libraries including scikit-learn, pycaret, matplotlib, and seaborn. 

\subsection{Interactive visualization for model evaluation and decision-making}

Visual dashboards can serve as a helpful tool for making the outputs of analytical models more comprehensible to stakeholders by converting predictive analytics into easily interpretable visual formats. This transformation is particularly beneficial for users from various backgrounds, including farmers, policymakers, and researchers, as it facilitates their engagement with and comprehension of the data. For example, platforms like Streamlit and PowerBI have been used to build visual web applications and dashboards across diverse fields, such as bioinformatics, bacterial testing, financial auditing, twitter sentiment analysis, credit card fraud detection, drug target prioritization, and pharmaceutical sales forecasting \cite{jain2022credit, nantasenamat2023building, patil2022live, belghith2023new, abusager2020using, nickell2023introductory}. 

In this study, we utilize the dashboard to not only visualize the model's predictions, such as forecasted yields in each Indian district, but also to offer insights into the model's diagnostic aspects, like its varying predictive accuracy across different regions. We developed two distinct visual dashboards: the first provides a clear, spatially detailed overview of the model's forecasts, and the second aids in identifying and understanding any potential errors in the model's performance.

\section{Results}

This study assessed the feasibility of predicting pre-harvest rice yields in India using machine learning models, with a focus on satellite and climate reanalysis data. The results, which are detailed in the subsequent sections and compared to other global studies of rice yield prediction, show that the models achieve strong performance across several metrics in out-of-sample tests. Such results affirm the potential of these models for rice yield forecasting and provide a benchmark for predictive precision in the field. 

\subsection{Overview of out-of-sample model performance}

Table~\ref{Table 2} summarizes the out-of-sample (validation set) performance across the models tested: R2, MAE, and MAPE values of up to 0.82, 0.29, and 0.16 respectively were achieved. Compared to out-of-sample results reported in previous literature on rice yield prediction in different parts of the world, the models perform well. For instance, one study which developed rice yield prediction models for China based on support vector machine regression, neural networks, and random forest, achieved R2 values ranging from 0.24 to 0.31 and MAE values ranging from 0.58 to 0/66 t/ha \cite{guo2021integrated}. Another study estimating rice yields in Vietnam’s Mekong Delta reported out-of-sample MAE values ranging from 0.46 to 0.55 t/ha for Winter and Summer rice models \cite{clauss2018estimating}. A study on county-level rice yield prediction in China’s Jiangsu province reported out-of-sample R2 values of 0.39 to 0.59 on an independent holdout set \cite{yu2023improved}. A study on pixel-scale rice yield prediction in South Korea reported test-set R2 values of 0.80 \cite{jeong2022predicting}. One image-driven yield prediction study reported test-set R2 values of 0.65 \cite{han2022rice}. A study using multi-temporal UAV-based multispectral vegetation indices reported test set R2 values of up to 0.80 \cite{su2023grain}. 

\begin{table}[h!]
\footnotesize
\caption{Model performance on out-of-sample test data shows that the top three (based on R2 and MAPE) models include Random Forest, CatBoost, and Light Gradient Boosting. Experiment 1 (“all features”) shows results for model runs including climate/satellite-derived features and the “Year” and “District” features. Experiment 2 shows results for model runs trained exclusively using climate and satellite data observations (“EO features only”).}
\vspace{10pt} 
\label{Table 2}
\centering
\begin{tabular}{lcccccccc}
\toprule
\textbf{Model} & \multicolumn{4}{c}{\textbf{Experiment 1 – all features}} & \multicolumn{4}{c}{\textbf{Experiment 2 – EO features only}} \\
\cmidrule(lr){2-5} \cmidrule(lr){6-9}
 & \textbf{MAE} & \textbf{RMSE} & \textbf{R2} & \textbf{MAPE} & \textbf{MAE} & \textbf{RMSE} & \textbf{R2} & \textbf{MAPE} \\
\midrule
Random Forest Regressor & 0.31 & 0.41 & 0.80 & 0.16 & 0.45 & 0.56 & 0.63 & 0.25 \\
CatBoost Regressor & 0.29 & 0.39 & 0.82 & 0.18 & 0.43 & 0.53 & 0.67 & 0.24 \\
Light Gradient Boosting Machine & 0.31 & 0.41 & 0.80 & 0.19 & 0.44 & 0.56 & 0.64 & 0.25 \\
Extreme Gradient Boosting & 0.33 & 0.43 & 0.78 & 0.19 & 0.44 & 0.58 & 0.61 & 0.23 \\
Orthogonal Matching Pursuit & 0.33 & 0.46 & 0.76 & 0.20 & 0.76 & 0.96 & -0.08 & 0.49 \\
Decision Tree Regressor & 0.41 & 0.56 & 0.63 & 0.20 & 0.58 & 0.81 & 0.24 & 0.33 \\
Bayesian Ridge & 0.33 & 0.46 & 0.76 & 0.21 & 7.51 & 11.81 & -161.70 & 4.15 \\
Gradient Boosting Regressor & 0.32 & 0.41 & 0.80 & 0.21 & 0.50 & 0.62 & 0.55 & 0.28 \\
Ridge Regression & 0.34 & 0.47 & 0.75 & 0.21 & 0.61 & 0.78 & 0.30 & 0.37 \\
Huber Regressor & 0.33 & 0.46 & 0.75 & 0.21 & 0.75 & 0.95 & -0.06 & 0.49 \\
K Neighbors Regressor & 0.39 & 0.51 & 0.70 & 0.21 & 0.55 & 0.70 & 0.44 & 0.31 \\
Linear Regression & 0.36 & 0.48 & 0.73 & 0.21 & 0.61 & 0.78 & 0.29 & 0.36 \\
AdaBoost Regressor & 0.45 & 0.55 & 0.65 & 0.27 & 0.63 & 0.75 & 0.34 & 0.40 \\
Passive Aggressive Regressor & 0.48 & 0.62 & 0.56 & 0.31 & 0.70 & 0.91 & 0.04 & 0.51 \\
Elastic Net & 0.66 & 0.81 & 0.24 & 0.41 & 0.74 & 0.94 & -0.04 & 0.49 \\
Lasso Regression & 0.80 & 0.99 & -0.13 & 0.53 & 0.74 & 0.94 & -0.03 & 0.49 \\
Lasso Least Angle Regression & 0.80 & 0.99 & -0.13 & 0.53 & 0.76 & 0.96 & -0.07 & 0.48 \\
Dummy Regressor & 0.80 & 0.99 & -0.13 & 0.53 & 0.80 & 0.99 & -0.13 & 0.53 \\
Least Angle Regression & 1.90 & 2.35 & -5.42 & 0.99 & 2.33 & 2.98 & -9.34 & 1.21 \\
\bottomrule
\end{tabular}
\end{table}

The results also perform well compared to studies which only reported in-sample performance metrics. One study on rice yield modelling in Bangladesh reported in-sample R2 values ranging from 0.44 to 0.91; out-of-sample performance was not reported \cite{islam2021development}. Another study on rice yield in China report in-sample R2 values of 0.77, lower than the out-of-sample R2 performance achieved in this study of 0.82 \cite{zhou2017predicting}. For rice yield prediction in the Philippines, one study reported an in-sample RMSE of 0.46 t/ha \cite{setiyono2018spatial}. Another study using drones reported in-sample R2 values of 0.60 to 0.81 for rice yield prediction in Japan based on NDVI \cite{guan2019assessing}.

\subsection{Errors, residuals, and SHAP value analysis}

\begin{figure}[!t]
    \centering
    \includegraphics[width=1\linewidth]{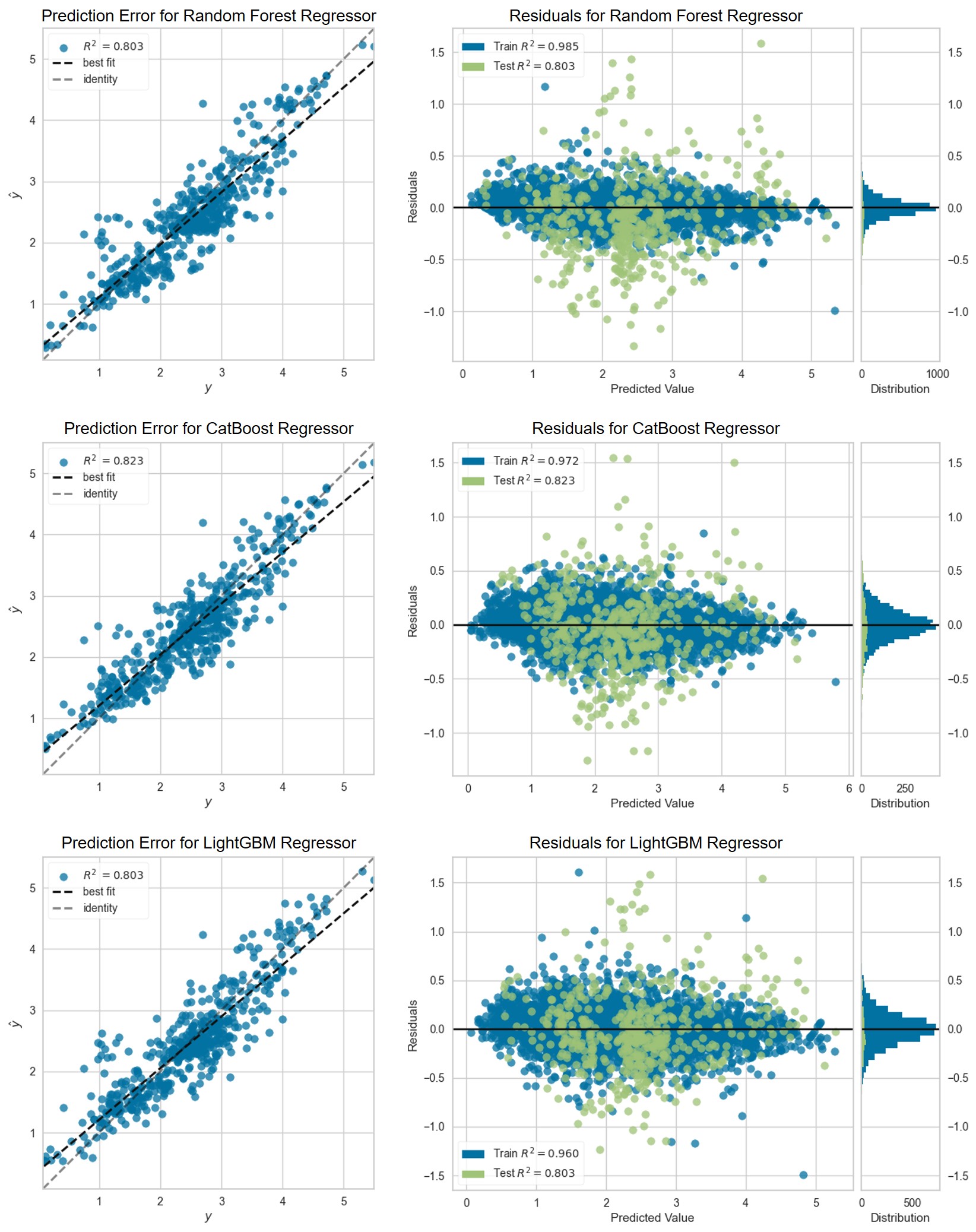}
    \caption{Comparative evaluation of selected models, including Random Forest, CatBoost, and LightGBM regressors using prediction error and residuals analysis. Two years of observations (502 observations in total) where used for the out-of-sample validation data, on which the Random Forest, CatBoost, and LightGBM models have test R² values of 0.80, 0.82, and 0.80 respectively. Residuals are mostly centered around zero, but CatBoost shows a skewness in error distribution. The histogram of residuals indicates Random Forest and CatBoost have a tighter error distribution compared to LightGBM's broader range.}
    \label{fig:2}
\end{figure}

As shown in Figure~\ref{fig:2}, observed and simulated yields show a high level of agreement for some of the top-performing models including the random forest, CatBoost, and LightGBM regressors. In addition, the residual plot residual plots show that the majority of both training set and test set observations are randomly dispersed along the horizontal axis, indicating a reasonable low level of bias and homoscedasticity. The distribution of residuals here is roughly centred around zero but with some skewness, indicating the potential presence of outliers. 

\begin{figure}[!t]
    \centering
    \includegraphics[width=1\linewidth]{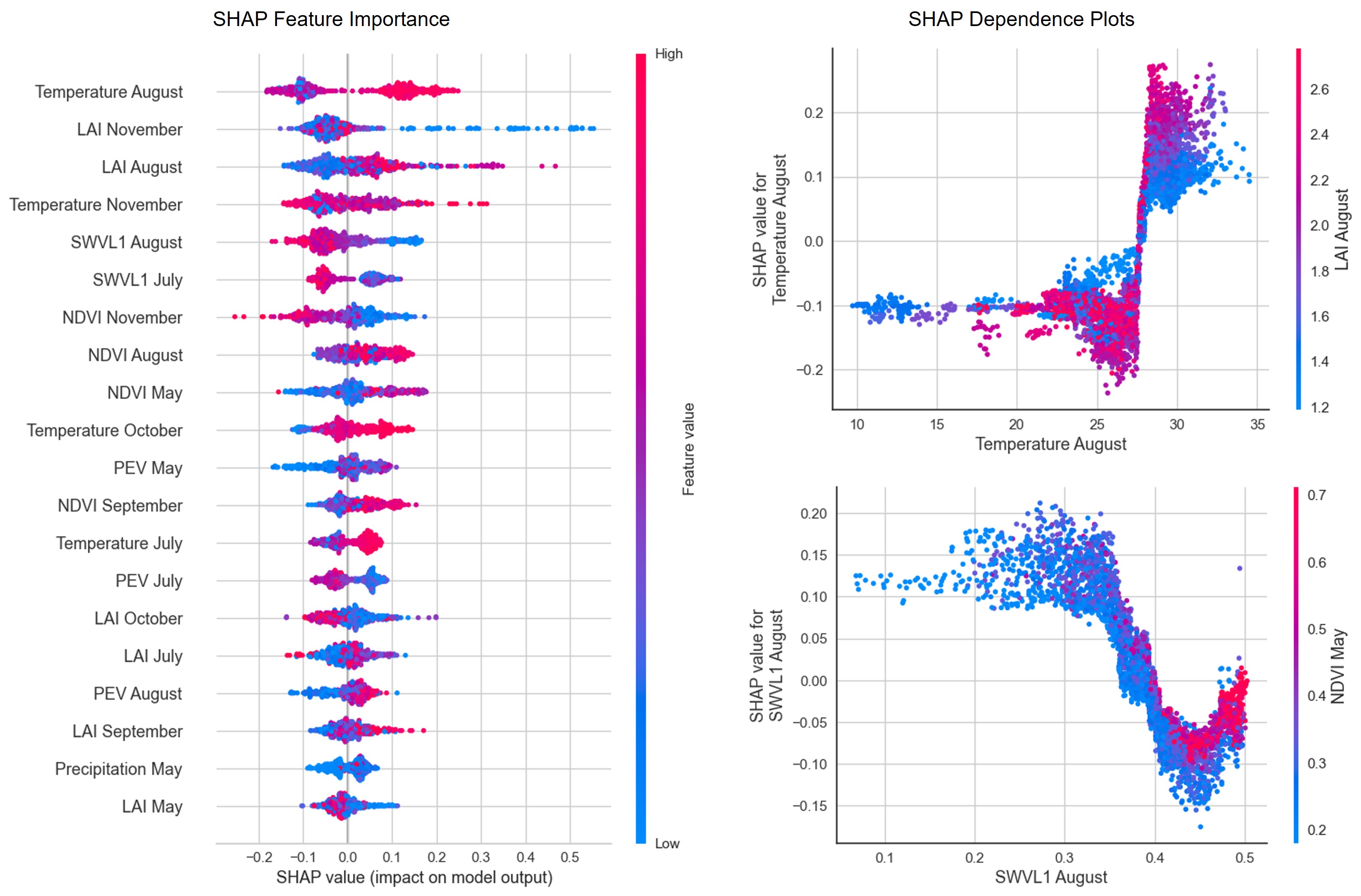}
    \caption{Interpretation of SHAP values for selected features in the rice yield prediction random forest model. The SHAP feature importance plot (left panel) exhibits the impact of various features on the model's output. Higher SHAP values indicate a greater influence on the predicted yield. The coloring on the feature importance plot represents the value of the feature for each data point. Blue points indicate low feature values, while pink points represent high feature values. This color gradient allows us to visualize not only the impact (magnitude of the SHAP value) each feature has on the model output but also the distribution of the feature's values. For instance, when examining 'Temperature August', we can see a mix of pink and blue points across a range of SHAP values, indicating a diverse range of temperatures in August within the dataset and how these varying temperatures correlate with the rice yield prediction. The top right panel presents a SHAP dependence plot for temperature in August, illustrating a correlation between higher temperatures and increased SHAP values for rice yield. The intensity of the color indicates the interaction effect, with a notable interaction with LAI in August, as higher LAI values (depicted in red) intensify the impact of temperature on yield. The bottom right panel depicts a SHAP dependence plot for soil water volume (SWVL1) in August, showing the relationship between SWVL1 values and SHAP values. This plot reveals that certain values of SWVL1 are associated with lower or higher SHAP values, indicating its varying influence on yield predictions, with the color intensity representing the interaction with NDVI in May.}
    \label{fig:3}
\end{figure}

In addition to the errors and residuals, the SHAP summary plots in Figure \ref{fig:3} concisely display the magnitude, prevalence and direction of a variable’s effect on final rice yield. The plots reveal that important variables include temperature, soil water volume (“SWVL1”), NDVI, and LAI in selected months. The importance ranking of these variables corroborates previous findings that that factors such as soil water content, temperature, and NDVI are important factors in estimating rice growth \cite{chakraborty2021assessing, jia2019effects}. 

A closer analysis of the specific impacts of features on rice yield is shown in the right-hand side panels of Figure \ref{fig:3}. For instance, increases in temperature in August, which coincides with the sowing to panicle initiation phases of rice growth, are associated with higher yields, corroborating previous findings in India that above average yields may be associated with higher maximum temperatures \cite{bal2023critical}. The full set of SHAP plots for all features is available in the analysis output on GitHub. 

\subsection{Interactive visualization tool}

\begin{figure}
    \centering
    \includegraphics[width=1\linewidth]{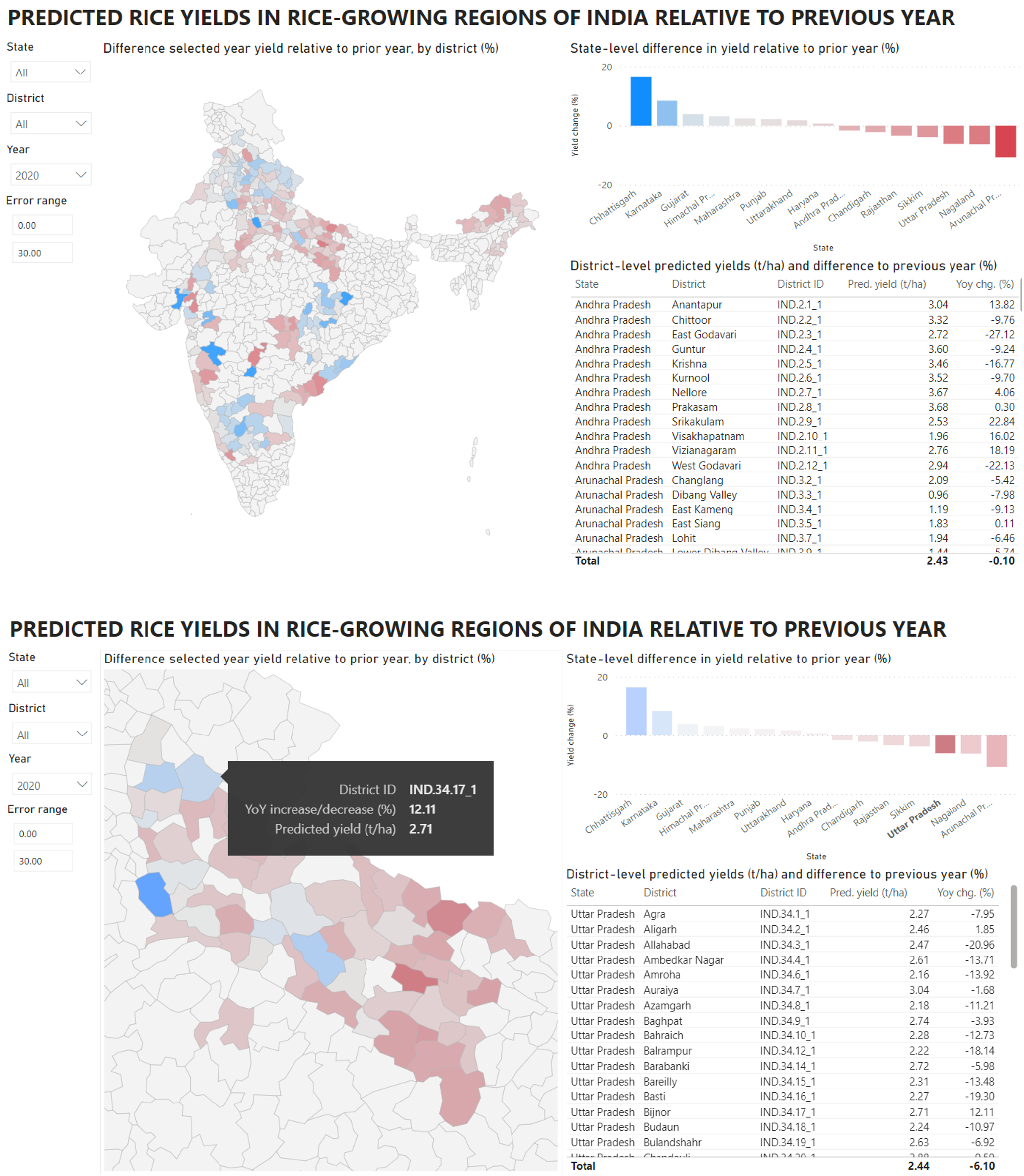}
    \caption{Interactive dashboard of yield prediction model outputs. The top panel shows a map of India with a colour coding applied to different districts to indicate the predicted yield values. Shades of blue indicate an increase in yield and shades of red denote a decrease in yield compared to the prior year's yield. Accompanying the map is a bar chart that provides a state-level summary and a table that enumerates the district-level predicted yields and the percentage change from the previous year across all states and districts. The bottom panel provides a similar comparative yield prediction, but focuses on the state of Uttar Pradesh.}
    \label{fig:4}
\end{figure}

\begin{figure}
    \centering
    \includegraphics[width=1\linewidth]{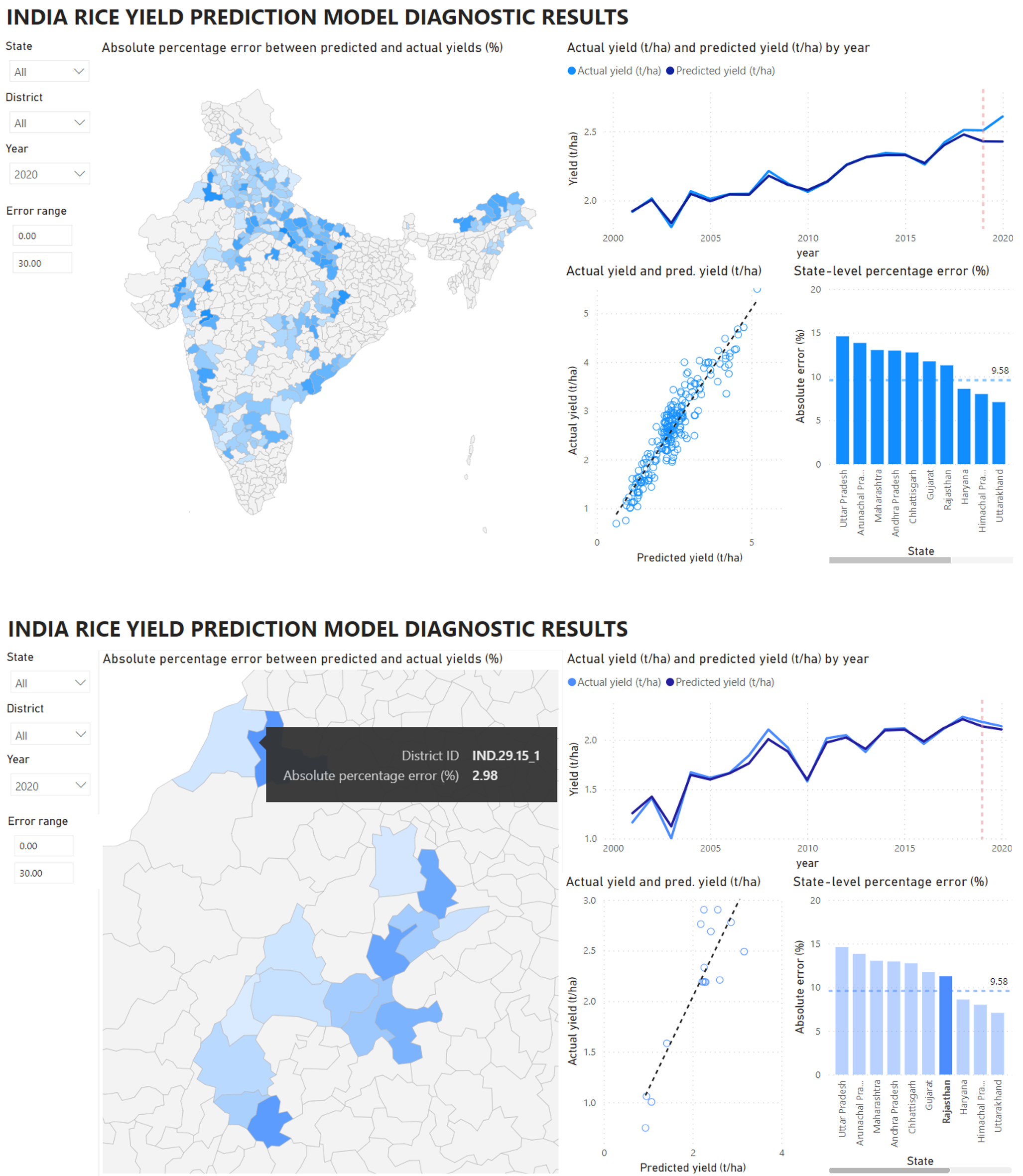}
    \caption{Interactive dashboard showing spatial view of yield prediction model error. The top panel provide model diagnostic information. The visual shows a map with the average percentage error by region; a scatter plot comparing the predicted yield and the actual yield; and a line graph showing the actual yield and predicted yield each year. The lower panel shows a similar view, zoomed in to districts within the state of Rajasthan.}
    \label{fig:5}
\end{figure}

In addition to the SHAP plots, which provide insight into how variables drive yield outcomes, two visual dashboards were developed to (a) provide an easy-to-understand, spatially explicit summary of model predictions, and (b) to help to identify potential biases in model performance. These are shown in Figure \ref{fig:4} and Figure \ref{fig:5}.

For example, in Figure \ref{fig:4} for the year 2020 (one of the test set years) the dashboard shows that districts in the state of Chhattisgarh with expected yield increases relative to the previous year included Jashpur, Korba, and Koriya, where yields were expected to increase by 52\%, 22\%, and 22\%, respectively. In another example, districts in the state of Gujarat such as Kheda and Sabar Kantha were expected to see yields decrease by 27\% and 21\% respectively according the model. 

In addition to a visual representation of the predictions, the dashboard also provides a spatial view of prediction errors in order to more easily identify areas where model predictions may be inaccurate (Figure \ref{fig:5}). For instance, the dashboard shows that on average, the absolute percentage error for predicted versus actual yields in ranged from an average of 7.1\% in districts in Uttarakhand to an average if 14.7\% in Uttar Pradesh, implying that the model may perform better in some regions than others.

\subsection{Limitations and future research directions}

Below, we present a non-exhaustive list of potential future research directions to build on the results of this research, including: refining model accuracy by exploring additional variables; communicating the outputs of yield prediction models in early warning systems by leveraging large language models; modeling how policymakers can help to disseminate the yield predictions-based early warning tools; and combining yield prediction modeling results with optimization approaches to support to anticipatory aid allocation efforts. 

\mpar
{\bf Modeling enhancements to increase predictive power.} Increased predictive power of rice yield model may potentially be achieved by incorporating a wider array of agronomically relevant climatological variables. Prior investigations have highlighted the influence of thermal extremes, precipitation extremes, daytime humidity variations, and solar radiation on rice yields \cite{bal2023critical}. In addition, agricultural yield models – particularly those that are grounded in time series analysis – may stand to gain from incorporating additional auto-regressive elements, rolling averages, or cumulative indices, such as total sunshine hours or cumulative rainfall since rice sowing \cite{cerqueira2021vest}. Beyond remotely sensed variables, yield models may also benefit from the inclusion of variables related to socio-economic changes in farmer populations, cultural practices that affect rice cultivation, and market or policy-related shifts that provide incentives for farmers to cultivate their crop in different ways \cite{liang2023analysis, jongeneel2020estimating}.

However, the augmentation of the feature set should also be approached with caution, as an expanded variable set can inadvertently lead to over-fitting, introduce feature redundancy, and complicate model interpretability.  In scenarios where model clarity and understandability are important, a more conservative approach with a reduced set of features might be advisable to streamline the interpretive process, allowing for clearer insights and more straightforward decision-making \cite{hastie2009elements}. Approaches which balance model interpretability with overall accuracy may be more helpful to decision makers. 

\mpar
\textbf{Leveraging large language models to communicate yield prediction results.} Additional research is needed on how access to climate and yield anomaly information via user-tailored interfaces can help mitigate meteorological shocks for agricultural communities. Studies have shown that inadequate access to climate information in South Asia has been observed as a factor driving perceived losses in farming communities \cite{islam2017determinants}.

Large language models (LLMs) may have potential to enhance the utility and accessibility of crop yield prediction models. By integrating LLMs with rice crop yield models, the data these yield models generate can be transformed into concise reports disseminated through channels such as text message-based farmer advisories or agricultural extension services. LLMs are able to process both structured and unstructured data, such as summarizing tabular data  on yield predictions for various regions and translating these summaries into different local languages, thus increasing their accessibility and ease of understanding \cite{mann2023susie, zhang2023foundation}.

Moreover, LLMs may have the potential to offer personalized agronomic advice based on location-specific yield predictions. Their proven capability in handling expert-level tasks across various fields, including agronomy, suggests they could be effectively deployed in agricultural settings \cite{romera2023mathematical,silva2023gpt4}. For instance, existing digital solutions like KissanGPT, a chatbot designed to assist farmers with queries such as optimal fertilizer application, indicate the practical applications of LLMs in agriculture \cite{tzachor2023llm}. LLMs could help make rice yield prediction models more accessible to farmers through web platforms or mobile messaging applications, like WhatsApp, allowing for interaction in their native languages. Examples of mobile-based agricultural early warning systems have included wheat rust disease alerts in Ethiopia and weather alerts in Zimbabwe \cite{allen2023early}.

Yield prediction systems that leverage LLM technology to communicate outputs may also help bolster agricultural extension services. Bangladesh, for instance, has boosted support services for farmers, including enhanced information access and extension services, via the government's 'Info Sarkar' project, which aims to link government offices nationwide and has established over 4,500 internet-equipped Union Digital Centres (UDCs) to aid rural communities. Studies indicate these centers are being effectively utilized by educated youths, suggesting a potential for these individuals to lead in disseminating climate adaptation strategies to farmers \cite{islam2017determinants}. 

Such agricultural extension services could benefit from additional digital aid such as LLMs that are able to distill agronomic science in a manner that is understandable and actionable for farmers, especially in contexts where agronomic advice should be disseminated in a context-specific manner. Advisories could be targeted based on whether farmers socio-economic status, the size of their farms, the level of internet connectivity, household income, amongst other factors. Past research has shown that differences in farmers backgrounds can significantly affect adoption of agricultural technologies, suggesting that tailored advice may be an important prerequisite for helping farmers adapt their agronomic practices in the face of climate-related risk \cite{Samal2011}. 

\mpar
\textbf{Modelling the dissemination of yield prediction outputs.} Building yield prediction systems that are tailored to local needs may help to boost resilience in agricultural communities. However, good design alone may be insufficient for widespread adoption. There is also a need to anticipate how external factors (such as mobile network coverage or regional agroclimatic variation) might affect the diffusion of this technology in a society across time and space, and how policymakers can enhance and sustain adoption. 

The adoption of these systems could potentially be modeled using principles of the Diffusion of Innovation theory \cite{Rogers2010}. This theory provides a framework to understand how innovations are taken up in a population, highlighting the role of early adopters and the subsequent spread through social and communication networks. In the context of yield prediction systems, this might involve analyzing how the perceived advantages and compatibility with existing practices influence the rate of adoption among farmers or agricultural extension workers.

Furthermore, computational models of diffusion, such as the Bass diffusion model, can offer insights into the expected rate of technology uptake. These models provide a means to simulate and anticipate the adoption curve, taking into account various societal and technological factors \cite{Bass1969}. By understanding the likely trajectory of technology diffusion, policies can be tailored to support and accelerate adoption, ensuring that the benefits of yield prediction systems are maximized. Applications of Bass-like models are numerous in the literature, but have yet to be applied to agricultural early warning technologies. Studies have applied the methodology to forecasting the diffusion of innovations including novel foods such as edible insects in the Netherland, groundwater pumps in Pakistan, preterm birth screening technology, fuel cell vehicles in China, household high-speed internet products, solar water heaters in Brazil, box office performance of movies, distributed solar generation, and cell phones in rural Bangladesh \cite{Horvat2020, Siddiqi2018, Grimm2018, Xian2022, Lartey2020, daSilva2020, Zhang2022, Ratcliff2016, Abud2022}. 

\mpar
\textbf{Leveraging yield prediction models for anticipatory aid allocation.} Another potential area of research could involve investigating how to effectively integrate yield prediction algorithms into decision-making tools for the public sector to enhance the strategic allocation of humanitarian aid in response to weather-related challenges. For example, algorithms that produce anticipated rice yields in specific areas could be combined with optimization algorithms can assist government decisions on the anticipatory allocation of resources such as financial support, fertilizers, fungicides, and water resources among affected farming communities.

While examples of such optimization applied specifically to anticipatory aid allocation in agricultural settings are limited, there is a growing body of research on applications of operations research applied to humanitarian aid \cite{Gutjahr2016}. Examples of using optimization approaches (such as linear mixed integer programming, stochastic programming, or multi-objective optimization) in relevant contexts include: aid disbursement following the 2010 Haiti earthquake \cite{Vitoriano2011}; optimal aid disbursement in response to internal displacement in northwest Syria \cite{Ismail2021}; post-disaster distribution of essential humanitarian aid (medicine, food, and water) from temporary warehouses to points of demand in Peru \cite{AceroCondor2023}; optimal location planning of warehouse locations to store relief items in Thailand \cite{Boonmee2020}; logistics distribution of essential relief items during COVID-19 lockdowns in Bangladesh \cite{Adnan2022}; optimization of United Nations Humanitarian Response Depot distribution plans \cite{Eliguzel2023}; humanitarian relief logistics in both pre- and post-disaster situations in presence of uncertainty \cite{Ghasemi2022}; and the World Food Programme’s Optimus tool, which leverages linear programming to optimize food aid operations \cite{Peters2021, Peters2022}.

\section{Conclusion}

This study advances the state of the art in district-level rice yield prediction in India through an integrated approach, combining ERA5 climate reanalysis, MODIS satellite vegetation indices, and a novel, spatially matched yield dataset. By evaluating 19 machine learning models, the research establishes benchmarks for accuracy, achieving out-of-sample R2, MAE, and MAPE values of up to 0.82, 0.29, and 0.16, respectively. The development of an interactive dashboard tool offers a means for visualizing yield predictions and assessing model performance across regions. This approach not only demonstrates the feasibility of using machine learning for rice yield forecasting at the district level in India, but also provides a benchmark for predictive accuracy against which further algorithmic innovations can be easily compared. The results offer evidence that machine learning-based rice yield prediction may have the potential to augment Indian agricultural early warning systems with robust crop yield prediction capabilities.


\newpage

\bibliographystyle{unsrt}
\bibliography{bibliography}

\end{document}